\documentclass[twoside,11pt]{article}
\pdfoutput=1 

\usepackage{jmlr2e}
\usepackage{algorithm2e}
\usepackage{booktabs}
\usepackage{multirow}
\jmlrheading{x}{201x}{x-xx}{x/xx}{xx/xx}{author list}

\ShortHeadings{autoBagging: Learning to Rank Bagging Workflows with Metalearning}{F\'{a}bio Pinto, V\'{i}tor Cerqueira, Carlos Soares and Jo\~{a}o Mendes-Moreira}
\firstpageno{1}

\begin{document}

\title{autoBagging: Learning to Rank Bagging Workflows with Metalearning}

\author{\name F\'{a}bio Pinto \email fhpinto@inesctec.pt \\
			 \name V\'{i}tor Cerqueira \email vmac@inesctec.pt \\
			 \name Carlos Soares \email csoares@fe.up.pt\\
			 \name Jo\~{a}o Mendes-Moreira \email jmoreira@fe.up.pt\\
      \addr INESC TEC\\
      Faculty of Engineering, University of Porto\\
      Rua Dr. Roberto Frias, s/n\\
      Porto, Portugal 4200-465
}

\editor{ Editor}

\maketitle

\begin{abstract}
Machine Learning (ML) has been successfully applied to a wide range of domains and applications. One of the techniques behind most of these successful applications is Ensemble Learning (EL), the field of ML that gave birth to methods such as Random Forests or Boosting. The complexity of applying these techniques together with the market scarcity on ML experts, has created the need for systems that enable a fast and easy drop-in replacement for ML libraries. Automated machine learning (autoML) is the field of ML that attempts to answers these needs. Typically, these systems rely on optimization techniques such as bayesian optimization to lead the search for the best model. Our approach differs from these systems by making use of the most recent advances on metalearning and a learning to rank approach to learn from metadata. We propose \textit{autoBagging}, an autoML system that automatically ranks 63 bagging workflows by exploiting past performance and dataset characterization.  Results on 140 classification datasets from the OpenML platform show that \textit{autoBagging} can yield better performance than the Average Rank method and achieve results that are not statistically different from an ideal model that systematically selects the best workflow for each dataset. For the purpose of reproducibility and generalizability, \textit{autoBagging} is publicly available as an R package on CRAN.
\end{abstract}


\section{Introduction}
\label{sec:intro}

Ensemble learning (EL) has proven itself as one of the most powerful techniques in Machine Learning (ML), leading to state-of-the-art results across several domains~\citep{fernandez2014we}. Methods such as bagging, boosting or Random Forests are considered some of the favourite algorithms among data science practitioners. However, getting the most out of these techniques still requires significant expertise and it is often a complex and time consuming task. Furthermore, since the number of ML applications is growing exponentially, there is a need for tools that boost the data scientist's productivity~\citep{mediumArticle}.

The resulting research field that aims to answers these needs is Automated Machine Learning (autoML). It is a field that merges ideas and techniques from several ML and optimization topics, such as Bayesian optimization, metalearning (MtL) and algorithm selection. In the past few years it was possible to assess important innovations in the field, which enable data science practitioners, including non-experts, to efficiently create fine-tuned predictive models with minimum intervention.

In this paper we address the problem of how to automatically tune an EL algorithm, covering all components within it: generation (how to generate the models and how many), pruning (which technique should be used to prune the ensemble and how many models should be discarded) and integration (which model(s) should be selected and combined for each prediction). We focus specifically in the bagging algorithm~\citep{breiman1996bagging} and four components of the algorithm: 1) the number of models that should be generated 2) the pruning method 3) how much models should be pruned and 4) which dynamic integration method should be used. For the remaining of this paper, we call to a set of these four elements a bagging workflow.

Our proposal is \textit{autoBagging}, a system that combines a learning to rank approach together with metalearning to tackle the problem of automatically generate bagging workflows. Ranking is a common task in information retrieval. For instance, to answer the query of a user, a search engine ranks a plethora of documents according to their relevance. In this case, the query is replaced by new dataset and \textit{autoBagging} acts as ranking engine. 

\begin{figure}[h]
	\centering
	\includegraphics[scale=0.45]{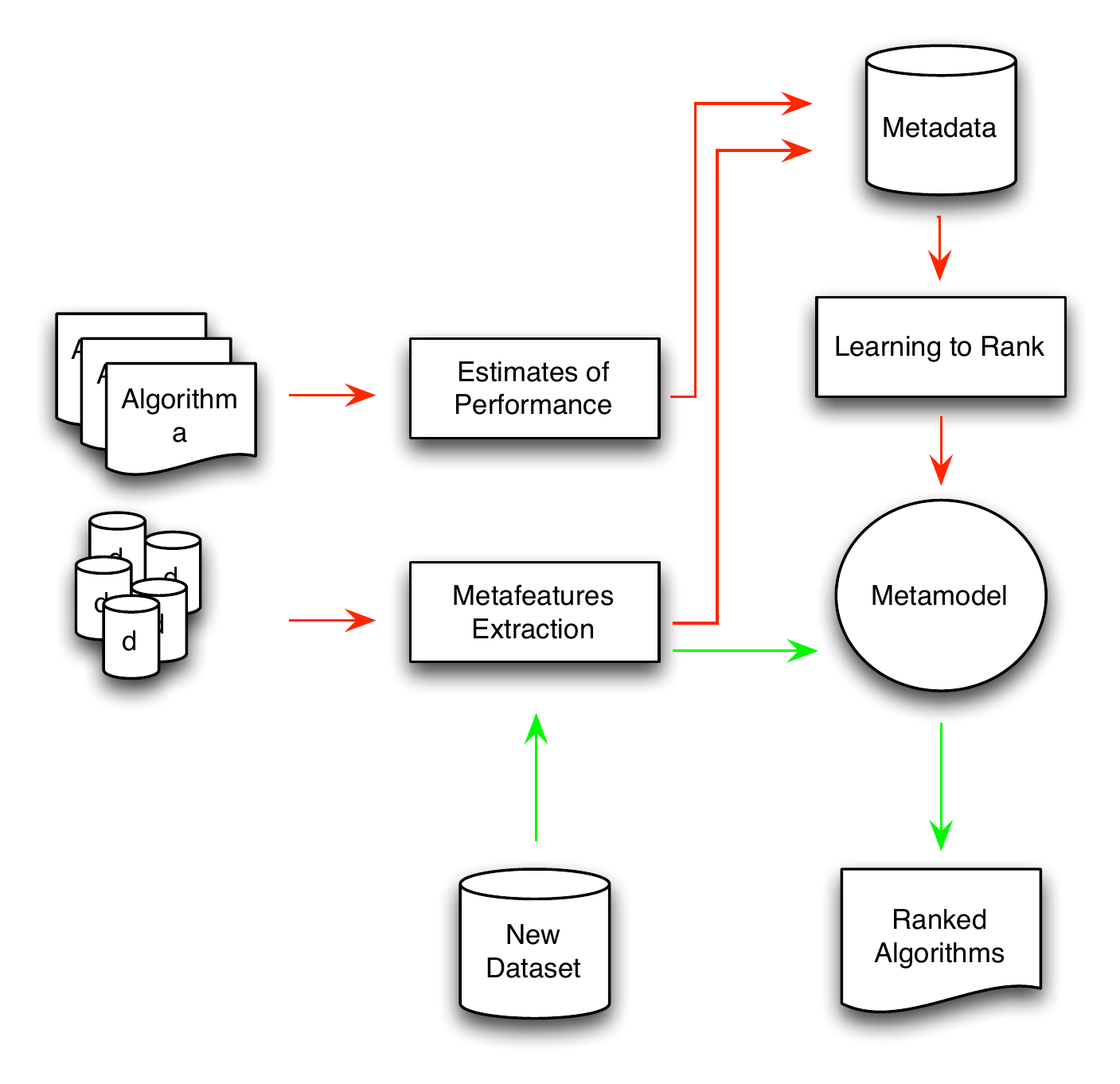}
	\caption{Learning to Rank with Metalearning. The red lines represent offline tasks and the green ones represent online ones.}
	\label{scheme}
\end{figure}

Figure~\ref{scheme} shows an overall schema of the proposed system. We leverage the historical predictive performance of each workflow in several datasets, where each dataset is characterised by a set of metafeatures. This metadata is then used to generate a metamodel, using a learning to rank approach. Given a new dataset, we able to collect metafeatures from it and feed them to the metamodel. Finally, the metamodel outputs an ordered list of the workflows, specifically tuned to the characteristics of the new dataset. 

We tested the approach in 140 classification datasets from the OpenML platform for collaborative ML~\citep{vanschoren2014openml} and 63 bagging workflows, that include two pruning techniques and two dynamic selection techniques. We give details on these workflows in Section~\ref{sec:ab}. Results show that \textit{autoBagging} has a better performance than two strong baselines, Bagging with 100 trees and the average rank. Furthermore, testing the top 5 workflows recommended by \textit{autoBagging} guarantees an outcome that is not statistically different from the Oracle, an idealistic method that for each dataset always selects the best workflow.

This paper is organized as follows. Section~\ref{sec:rw} describes the state-of-the-art regarding autoML and metalearning, with particular emphasis for the approaches more similar to ours. In Section~\ref{bagging_workflows} we introduce the concept of bagging workflows and describe the components from each they are designed with. Section~\ref{sec:ab} presents AutoBagging from a more formal perspective. Section~\ref{sec:exper} presents the experiments carried to evaluate our approach. Finally, Section~\ref{sec:cfw} concludes the paper and sets directions for future work.

For the purpose of reproducibility and generalizability, \textit{autoBagging} is publicly available as an R package.~\footnote{https://github.com/fhpinto/autoBagging}

\section{Related Work}
\label{sec:rw}

In this Section we provide a brief overview of systems designed with intent of generating automatic recommendations or ranking of ML algorithms or workflows. After a careful analysis of the state-of-the-art, we split it into three categories: 1) systems that use only metalearning approach, without any kind of optimization component; 2) systems that make use of optimization procedures, such as bayesian optimization and 3) systems that leverage metalearning and optimization procedures.

Finally, in the last sub-section, we discuss how some autoML systems recommend ensemble learning algorithms to the user and how our approach differs from previous ones regarding this feature.

\subsection{Metalearning based}

The first automated framework proposed to support machine learning processes was the Data Mining Advisor (DMA)~\citep{giraud2005data}. The system used an instance-based learning approach to relate the performance of the learning algorithms with simple and statistical metafeatures computed from the datasets~\citep{brazdil2003ranking}. 

This line of research was later on overpowered by the characterization of datasets through landmarkers~\citep{pfahringer2000tell}, such as learning curves~\citep{leite2005predicting} or pairwise meta-rules~\cite{sun2013pairwise}.

\subsection{Optimization based}

Evaluating ML algorithms and/or ML workflows is typically very time consuming and computationally expensive. In practice, it is not feasible to evaluate all learning algorithms with 10-fold cross validation for a given dataset (particularly if the dataset is of high dimensionality) and choose the one that minimizes the error measure. Therefore, researchers have been developing search procedures and optimization algorithms that can in fact do this in reasonable time.

Bayesian optimization is the field within optimization that has had the most success carrying out these type of tasks. One of the algorithms that is responsible for this success is SMAC~\citep{hutter2011sequential}, an approach that constructs explicit regression models to describe the relationship between the target algorithm performance and the hyperparameters. The ability of SMAC to deal both categorical and continuous hyperparameters is one of the reasons behind its success.

The development of bayesian optimization algorithms for algorithm configuration has led to the emergence of systems such as Auto-WEKA~\citep{thornton2013auto}, that makes use of SMAC and the machine learning library WEKA to automatically generate workflows for classification datasets.

More recently, the Hyperband~\citep{li2016hyperband} method was proposed as an alternative for bayesian optimization algorithms. The method uses a pure-exploration algorithm for multi-armed bandits to exploit the iterative algorithms of machine learning. Essentially, the authors approach the automatic model \textit{selection} problem as an automatic model \textit{evaluation} problem. By exploring this perspective, they report better results than the bayesian optimization methods.

\subsection{Optimization plus Metalearning}

Some autoML systems combine bayesian optimization with metalearning, particularly useful to act as a warm start for the optimization procedure. An example of such system is auto-sklearn~\citep{feurer2015efficient}. Given a new dataset, the system starts by comparing the characteristics of that dataset with past performance of ML workflows on similar datasets (using a set of simple, statistical and information-theoretic metafeatures and k-NN). After this warm start, the optimization procedure is carried out by SMAC. Finally, the system also has the ability to form ensembles from models evaluated during the optimization.

%~\citep{malkomes2016bayesian}

\subsection{Ensemble focused autoML}

Some attempts have been made in creating autoML systems that are able to provide suggestions of ensembles. Again, and probably the most notorious one, is auto-skelarn, as described above. Another proposal on this matter is made on~\citep{lacoste2014sequential}, where the authors optimize ensembles based on bootstrapping the validation sets to simulate multiple independent hyperparameter optimization processes and combined the results with the agnostic Bayesian combination method.

One of the problems with the two approaches described is that the generation of the ensemble is rather \textit{ad hoc}. That means, it does not take into account important properties that are known to affect the performance of ensembles. Specifically, complementarity between among models and the overall diversity of the ensemble. We argue that ensemble generation must take into account these concepts and we should avoid a simple averaging of predictions from several models. For instance, in~\citep{levesque2016bayesian}, the authors use Bayesian optimization directly to estimate which prediction model is the best candidate to be added to the ensemble. In this paper, we use as basis a well known and studied ensemble learning algorithm (bagging) and we generate ensembles that make use of several EL techniques proposed in the literature.

\section{Bagging Workflows}
\label{bagging_workflows}

The Ensemble Learning (EL) literature can be split into three main topics: ensemble generation, ensemble pruning and ensemble integration~\citep{mendes2012ensemble}. It can also be seen as a process of three phases: 1) generating an accurate and diverse set of models; 2) prune the ensemble in order to decrease its size and attempt to improve its generalization ability; and finally, 3) select a function to aggregate the predictions of each single model of the ensemble. This can be achieved by a static (that does not take into account the characteristics of the test instance, such as stacking) or dynamic method (that chooses different subsets of models according to the characteristics of the test instance).

Bagging, one of the most popular EL algorithms, can also be decomposed at the light of the structure that we described above~\citep{breiman1996bagging}. Generically, given a training data set, a sample with replacement (a bootstrap sample) of the training instances is generated. The process is repeated \textit{k} times and \textit{k} samples of the training instances are obtained. Then, from each sample, a model is generated by applying a learning algorithm. In terms of aggregating the outputs of the base learners and building the ensemble, typically, bagging uses two of the most common ones: voting for classification (the most voted label is the final prediction) and averaging for regression (the predictions of all the base learners are averaged to form the ensemble prediction).

In this paper, we introduce the concept of a bagging workflow, that can also be decomposed into three components: generation, pruning and integration. The following subsections describe some of the methods that can be used within each of these components.

\subsection{Generation}

As mentioned before, typically, bagging algorithms generate ensembles by applying a learning algorithm to bootstrap samples of the training data. However, there are some hyperparameters that can be taken into account to exploit the versatility of bagging, such as:
\begin{itemize}
	\item the sampling strategy. Although bootstrap sampling is by far the most common sampling strategy in bagging, there are also reports of interesting results using sub-sampling without replacement and sampling of random subspaces~\citep{ho1998random}.
	\item the learning algorithm used to generated the models. Decision trees and neural networks are among the favourite, given their unstable learning property~\citep{breiman1996bagging}.
	\item how many models to generate. On the seminal paper in which bagging was introduced~\citep{breiman1996bagging}, the author claimed that 50 or 100 single models should be enough to achieve good results. However, more recent studies showed that this problem is highly dataset dependent~\citep{hernandez2013large}.
\end{itemize}

\subsection{Pruning}

Given the widely spread use among data science practitioners, bagging is also one the most studied algorithms~\citep{bauer1999empirical}. One of the discoveries made by researchers is that an efficient pruning of a bagging ensemble could to a smaller ensemble size and also to generalization improvements~\citep{zhou2002ensembling}. This has led a stream of research focused specifically on pruning techniques for bagging ensembles. Since a detailed overview of these techniques would be out of scope of this paper, we refer the reader to some important papers in the field~\citep{martinez2009analysis, qian2015pareto}. Essentially, these techniques combine the concepts of accuracy and diversity in ensemble learning to search for a subset of models that guarantees the same performance of the full ensemble or even improves it. This search procedure is often led by some heuristic or an optimization algorithm.

Therefore, from the ensemble pruning phase of constructing bagging workflows, two hyperparameters must be considered:
\begin{itemize}
	\item the pruning method to be used.
	\item the percentage of models that should be pruned. Again, studies have shown that this a highly dataset dependent hyperparameter~\citep{hernandez2013large}.
\end{itemize}

\subsection{Integration}

As mentioned before, the method regarding ensemble integration are split into two groups: static and dynamic. In the former, the weights assigned to each model in the ensemble are a constant value; in the later, the weights vary according to the instance to be predicted. In the dynamic group, we distinguish between methods for selection (when a single model is selected) or combination of models (when more that one model can be selected).

Regarding static methods, the most well known is stacking~\citep{wolpert1992stacked}. Regarding dynamic methods, again, research has shown that this hyperparameter is highly problem dependent~\citep{britto2014dynamic}. A large empirical comparison of these techniques can be found in~\citep{pinto2016chade}. A full description of these techniques is out of scope of this paper so we refer the reader to the original papers.

\section{autoBagging: Learning to Rank Bagging Workflows}
\label{sec:ab}

In this Section we present \textit{autoBagging}. Although for this paper we focused on providing ranking of bagging workflows, we believe that the approach is generic for the algorithm/workflow ranking in ML. Therefore, we describe the method from a generic perspective and we provide more specific details on the application to bagging workflows in Section~\ref{sec:exper}.

We recall Figure~\ref{scheme} for a brief overview of the method. We start by describing the learning approach and then how we collected the metadata, both metafeatures and metatarget, to be able to learn at the meta-level.

\subsection{Learning Approach} 

We approach the problem of algorithm selection as a learning to rank problem~\citep{liu2009learning}. Lets take $\mathcal{D}$ as the dataset set and $\mathcal{A}$ as the algorithm set. $\mathcal{Y} = \{1,2,...,l\}$ is the label set, where each value represents a relevance score, which represents the relative performance of a given algorithm. Therefore, $l \prec l-1 \prec ... \prec 1$, where $\prec$ represents an order relationship. 

Furthermore, $D_{m} = \{d_{1}, d_{2}, ..., d_{m}\}$ is the set of datasets for training and $d_{i}$ is the $i$-th dataset, $A_{i} = \{a_{i,1}, a_{i,2}, ..., a_{i,n_{i}}\}$ is the set of algorithms associated with dataset $d_{i}$ and $\textbf{y}_{i} = \{y_{i,1}, y_{i,2}, ..., y_{i,n_{i}}\}$ is the set of labels associated with dataset $d_{i}$, where $n_{i}$ represents the sizes of $A_{i}$ and $\textbf{y}_{i}$; $a_{i,j}$ represents the $j$-th algorithm in $A_{i}$; and $y_{i,j} \in Y$ represents the $j$-th label in $\textbf{y}_{i}$, representing the relevance score of $a_{i,j}$ with respect to $d_{i}$. Finally, the meta-dataset is denoted as $S = \{(d_{i}, A_{i}), \textbf{y}_{i}\}_{i=1}^{m}$.

We use metalearning to generate the metafeature vectors $x_{i,j} = \phi(d_{i}, a_{i,j})$ for each dataset-algorithm pair, where $i = 1,2,...,m$; $j = 1,2,...,n_{i}$ and $\phi$ represents the metafeatures extraction functions. These metafeatures can describe $d_{i}$, $a_{i,j}$ or even the relationship between both. Therefore, taking $\textbf{x}_i = \{x_{i,1}, x_{i,2}, ..., x_{i,n_{i}}\}$ we can represent the meta-dataset as $S' = \{(\textbf{x}_{i} , \textbf{y}_{i})\}_{i=1}^{m}$.

Our goal is to train a meta ranking model $f(d,a) = f(x)$ that is able to assign a relevance score to a given new dataset-algorithm pair $d$ and $a$, given $x$.

\subsection{Metafeatures}

We approach the problem of generating metafeatures to characterize $d$ and $a$ with the aid of a framework for systematic metafeatures generation~\citep{pinto2016towards}. Essentially, this framework regards a metafeature as a combination of three components: meta-function, a set of input objects and a post-processing function. The framework establishes how to systematically generate metafeatures from all possible combinations of object and post-processing alternatives that are compatible with a given meta-function. Thus, the development of metafeatures for a MtL approach simply consists of selecting a set of meta-functions (e.g. entropy, mutual information and correlation) and the framework systematically generates the set of metafeatures that represent all the information that can be obtained with those meta-functions from the data.

For this task in particular, we selected a set of meta-functions that are able to characterize the datasets as completely as possible (measuring information regarding the target variable, the categorical and numerical features, etc) the algorithms and the relationship between the datasets and the algorithms (who can be seen as landmarkers~\citep{pfahringer2000tell}). Therefore, the set of meta-functions used is:
\begin{itemize}
	\item Skewness
	\item Pearson's correlation
	\item Maximal Information Coefficient (MIC~\citep{reshef2011detecting})
	\item Entropy
	\item Mutual Information
	\item Eta squared (from ANOVA test)
	\item R value of class overlap~\citep{oh2011new}
	\item Rank of each algorithm~\citep{brazdil2008metalearning}
\end{itemize}

Each meta-function is used to systematically measure information from all possible combination of input objects available for this task. We defined the input objects available as:
\begin{itemize}
	\item discrete descriptive data of the datasets
	\item continuous descriptive data of the datasets
	\item discrete output data of the datasets
	\item five sets of predictions (discrete predicted data) for each dataset (naive bayes, decision tree with depth 1, 2 and 3, and majority class)
\end{itemize}

For instance, if we take the example of using Entropy as meta-function, it is possible to measure information in discrete descriptive data, discrete output data and discrete predicted data (if the base-level problem is a classification task). After computing the entropy of all these objects, it might be necessary to aggregate the information in order to keep the tabular form of the data. Take for the example the aggregation required for the entropy values computed for each discrete attribute. Therefore, we choose a palette of aggregation functions to capture several dimensions of these values and minimize the loss of information by aggregation. In that sense, the post-processing functions chosen were:
\begin{itemize}
	\item average
	\item maximum
	\item minimum
	\item standard deviation
	\item variance
	\item histogram binning
\end{itemize}

Given these meta-functions, the available input objects and post-processing functions, we are able to generate a set of 146 metafeatures. To this set we add eight metafeatures: the number of examples of the dataset, the number of attributes and the number of classes of the target variable; and five landmarkers (the ones already described above) estimated using accuracy as error measure. Furthermore, we add four metafeatures to describe the components of each workflow: the number of trees, the pruning method, the pruning cut point and the dynamic selection method. In total, \textit{autoBagging} uses a set of 158 metafeatures.

\subsection{Metatarget}

In order to be able to learn a ranking meta-model $f(d,a)$, we need to compute a metatarget that is able to assign a score $z$ to each dataset-algorithm pair $(d,a)$, so that:

\begin{equation}
\mathcal{F}: (\mathcal{D},\mathcal{A}) \rightarrow \mathcal{Z}
\end{equation}

where $\mathcal{F}$ is the ranking meta-models set and $\mathcal{Z}$ is the metatarget set.

To compute $z$, we use a cross validation error estimation methodology (we use a 4 fold cross validation in the experiments reported in this paper, Section~\ref{sec:exper}), in which we estimate the performance of each bagging workflow for each dataset using Cohen's kappa score~\citep{cohen1960coefficient}. On top of the estimated kappa score, for each dataset, we rank the bagging workflows. This ranking is the final form of the metatarget and it is then used for learning the meta-model.

\section{Experiments}
\label{sec:exper}

In this Section we describe the experiments performed to understand and evaluate \textit{autoBagging}. We also provide a brief exploratory analysis of the metadata collected from the experiments that are particularly interesting to understand some of the EL methods used.

\subsection{Experimental Setup}

Our experimental setup comprises 140 classification datasets extracted from the OpenML platform for collaborative machine learning~\citep{vanschoren2014openml}. We limited the datasets extracted to a maximum of 5000 instances, a minimum of 300 instances and a maximum of 1000 attributes, in order to speed up the experiments and exclude datasets that could be too small for some of bagging workflows that we wanted to test.

Regarding bagging workflows, taking into account all the hyperparameters described in Section~\ref{bagging_workflows} would result in a computational cost too large for our resources. Therefore, we limited the hyperparameters of the bagging workflows to four: number of models generated, pruning method, pruning cut point and dynamic selection method. Specifically, each hyperparameter could take the following values:

\begin{itemize}
	\item Number of models: 50, 100 or 200. Decision trees was chosen as learning algorithm.
	\item Pruning method: Margin Distance Minimization(MDSQ)~\citep{martinez2009analysis}, Boosting-Based Pruning (BB)~\citep{martinez2009analysis} or none.
	\item Pruning cut point: 25\%, 50\% or 75\%.
	\item Dynamic integration method: Overall Local Accuracy (OLA), a dynamic selection method~\citep{woods1997combination}; K-nearest-oracles-eliminate (KNORA-E)~\citep{ko2008dynamic}, a dynamic combination method; and none.
\end{itemize}

All the values of the hyperparameters described above generated 63 valid combinations. We tested these bagging workflows in the datasets extracted from OpenML with 4-fold cross validation, using Cohen's kappa as evaluation metric.

We used the XGBoost learning to rank implementation for gradient boosting of decision trees~\citep{Chen2016} to learn the metamodel as described in Section~\ref{sec:ab}. The decision tree implementation from this library has a very elegant way of dealing with missing values. Essentially, the tree splitting functionality assigns an instance with missing values to a default direction and then learns from the data the optimal default direction. This is particularly important for metalearning since the number of missing values is often quite high in these dataset (e.g., attribute correlation cannot be measured in a dataset without numeric attributes, which results in a missing value).

As baselines, at the base-level, we use 1) bagging with 100 decision trees 2) the average rank method, which basically is a model that always predicts the bagging workflow with best average rank in the meta training set and the 3) oracle, an ideal model that always selects the best bagging workflow for each dataset. As for the meta-level, we use as baseline the average rank method.

As evaluation methodology, we use an approach similar to the leave-one-out methodology. However, each test fold consists of all the algorithm-dataset pairs associated with the test dataset. The remaining examples are used for training purposes. The evaluation metric at the meta-level is the Mean Average Precision at 10 (MAP@10) and at the base-level, as mentioned before, we use Cohen's kappa. The methodology proposed by~\citep{demvsar2006statistical} was used for statistical validation of the results.

%We decided to not include time in the experiments since \textit{autoBagging} execution time only depends on the computation of metafeatures. Given the nature of these metafeatures, such as entropy or mutual information, the computation is extremely fast (no more than a minute for the largest datasets used in the experiments). However, we plan in future work to do a sensitivity analysis in which we analyse the scalability of the system by increasing the number of examples and/or features in the dataset.

For the purpose of reproducibility and generalizability, \textit{autoBagging} is publicly available as an R package.

\subsection{Exploratory Metadata Analysis}

Given the rich metadata collected from the experiments that we carried out, we proceed to draw some insights about the datasets and the workflows that we experimented with.

\begin{figure}[h]
	\centering
	\includegraphics[scale=0.5]{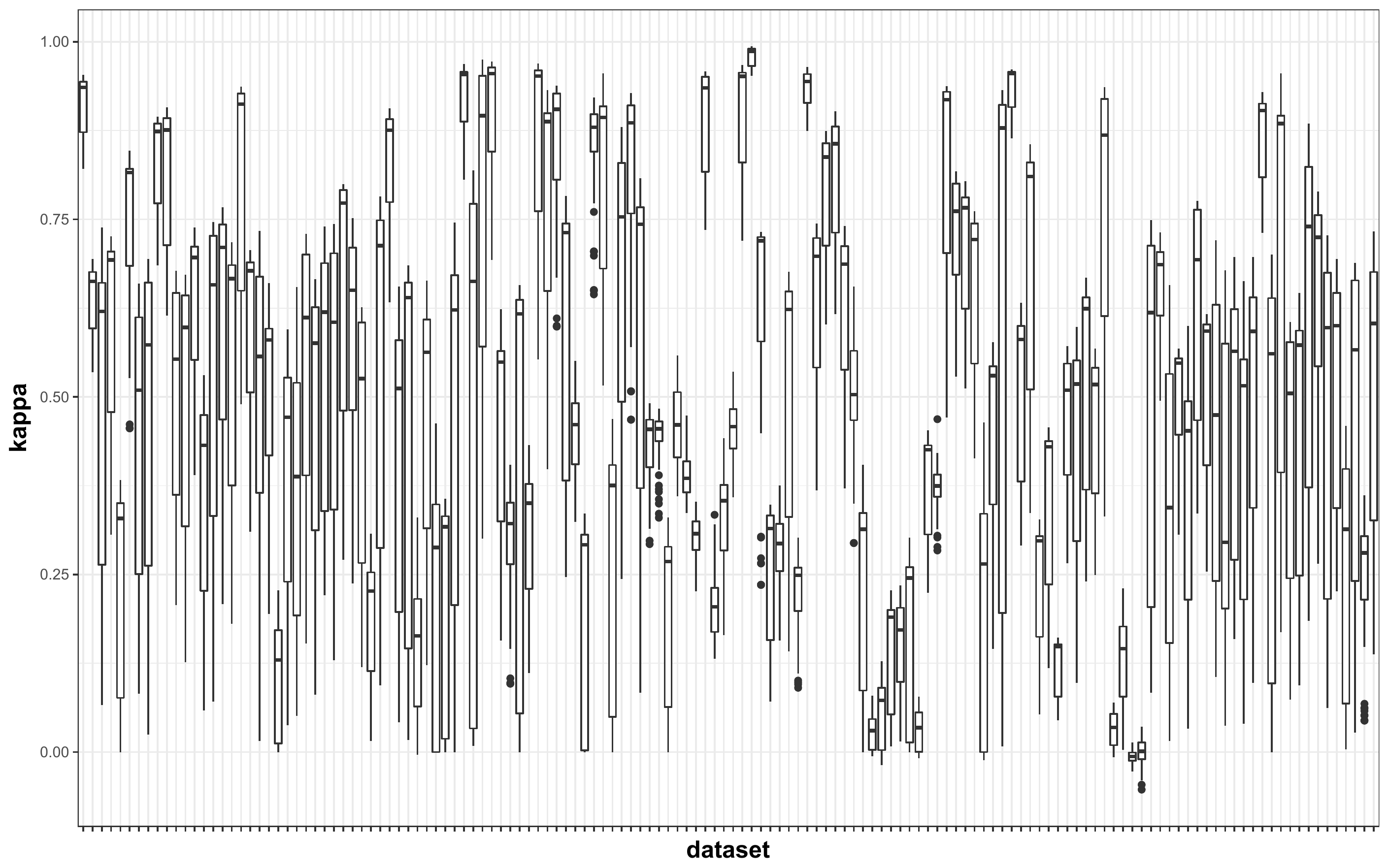}
	\caption{Boxlplots of the kappa values collected for each dataset from evaluating the performance of each bagging workflow.}
	\label{kappa_dataset}
\end{figure}

We can see by analysing Figure~\ref{kappa_dataset} that the range of kappa values for each dataset varies a lot. This is expected given the No Free Lunch theorem, that states that there is no one model that works best for every problem and "two algorithms are equivalent when their performance is averaged across all possible problems"~\citep{wolpert1996lack}. Even though all the models that we experimented with belong to the same family (bagging of decision trees), the pruning and dynamic integration components enable to generate very different predictive models. This is indicative that ranking these bagging workflows for each dataset is not an easy learning task.

\begin{figure}[h]
	\centering
	\includegraphics[scale=0.5]{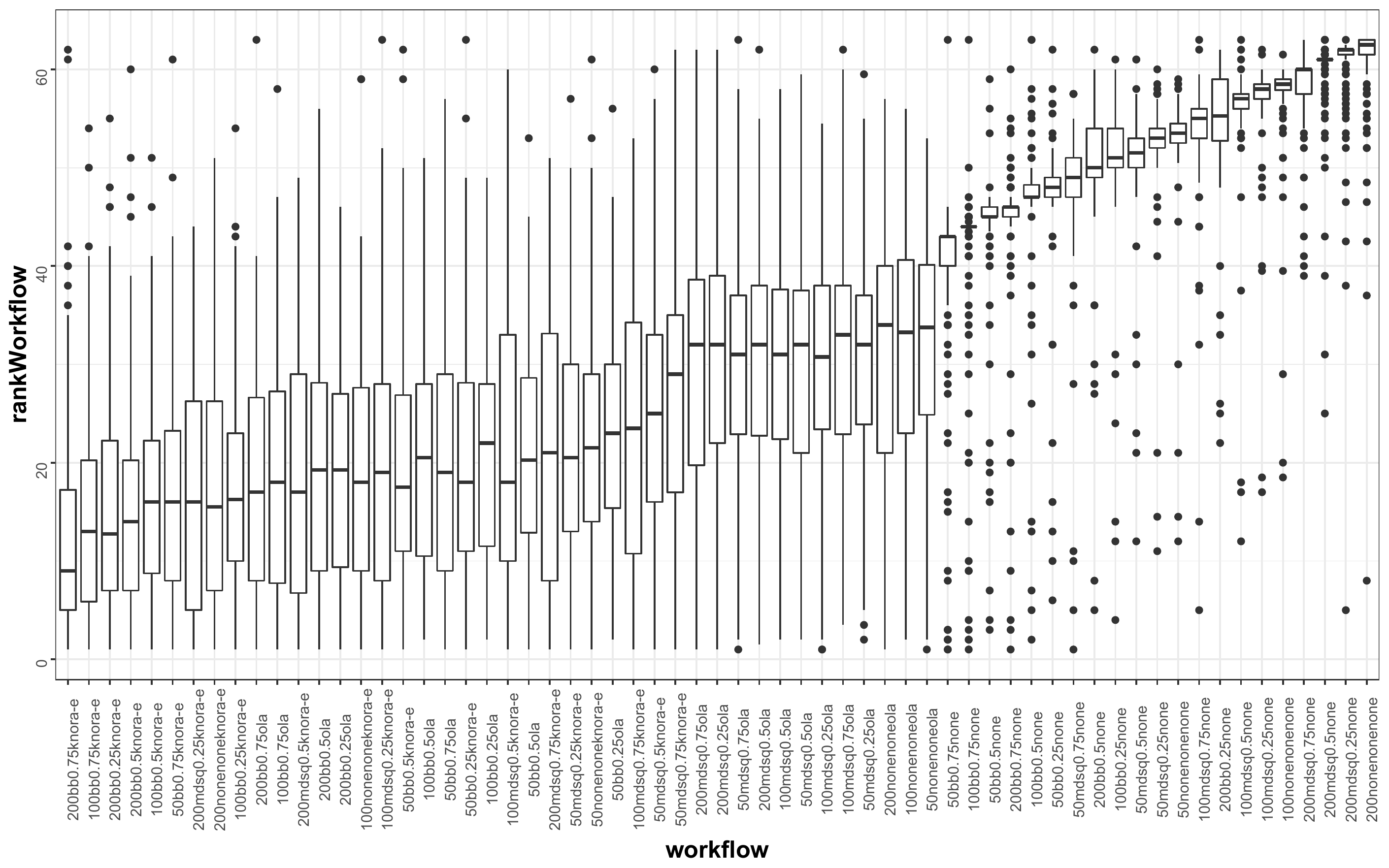}
	\caption{Boxlplots of the ranking scores collected for each bagging workflow. For instance, \textit{200bb0.75knora-e} represents a bagging workflow with 200 trees, to which boosting-based pruning is applied with a 75\% cut point and KNORA-E is used as dynamic integration technique.}
	\label{rank_workflow}
\end{figure}

Figure~\ref{rank_workflow} shows the boxplots of the ranking scores collected for each dataset, ordered by average ranking. We can take some insights about the bagging workflows performance from this graph:

\begin{itemize}
	\item on average, the bagging workflows that make use of BB pruning and KNORA-E as dynamic integration method seem to achieve better results
	\item in terms of pruning cut point, it seems that BB pruning works better with a large pruning cut point (e.g., 75\%) than MDSQ
	\item the bagging workflows that do not make use of any kind of dynamic integration method are worse on average than the ones that do
	\item both the top and the worst bagging workflows  are outliers for some dataset in terms of performance
\end{itemize}

\subsection{Results}

\begin{figure}[h]
	\centering
	\includegraphics[scale=0.5]{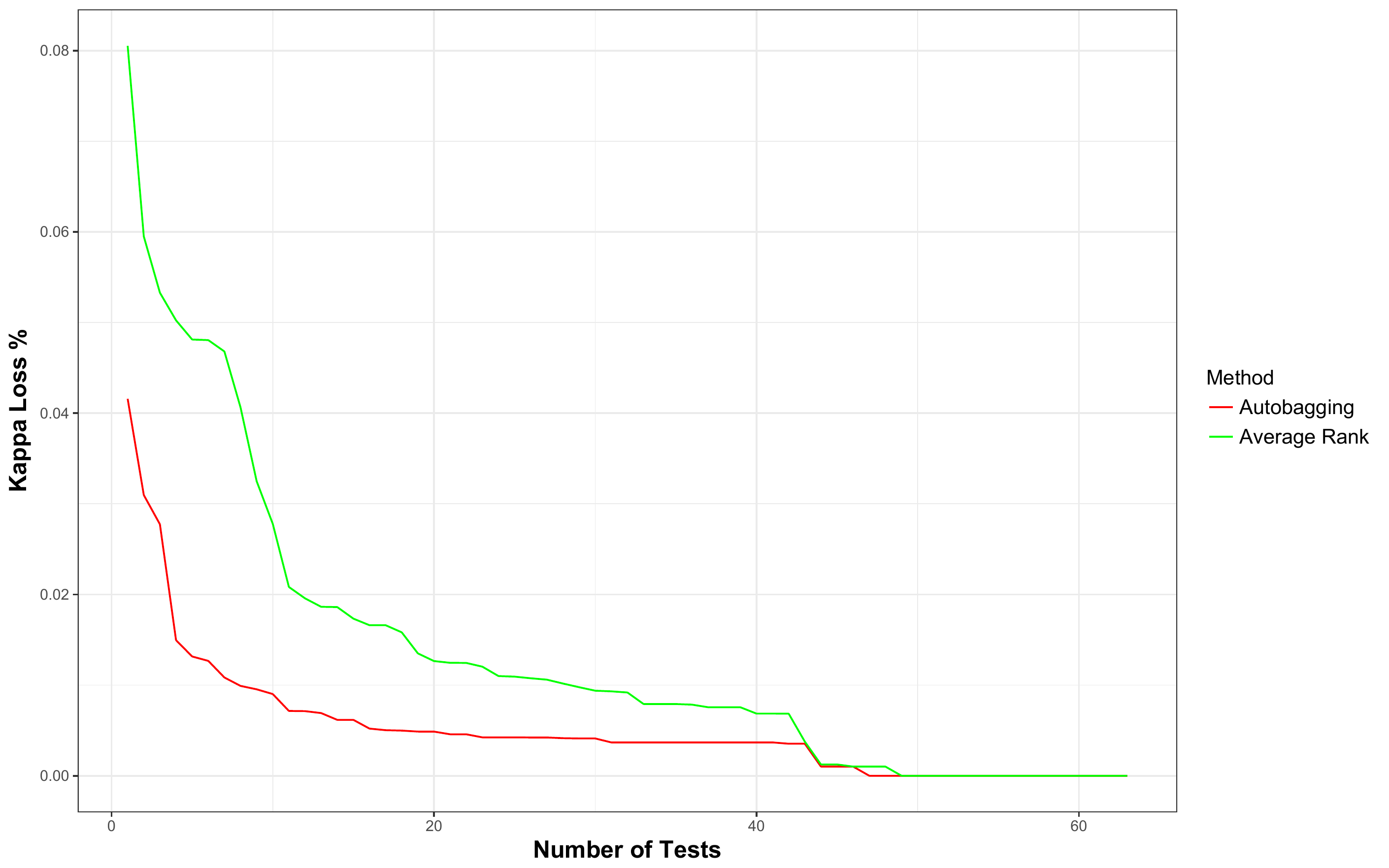}
	\caption{Loss curve comparing autoBagging with the Average Rank baseline.}
	\label{loss_curve}
\end{figure}

Figure~\ref{loss_curve} shows a loss curve, relating the average loss in terms of performance with the number of workflows tested following the ranking suggested by each method. The loss is calculated as the difference between the performance of the best algorithm ranked by the method in comparison with the ground truth ranking. The loss for all datasets is then averaged for aggregation purposes. We can see, as expected, that the average loss decreases for both methods as the number of workflows tested increases.

In terms of comparison between \textit{autoBagging} and the Average Rank method, it is possible to visualize that \textit{autoBagging} shows a superior performance for all the values of the $x$ axis. Interestingly, this result is particularly noticeable in the first tests. For instance, if we test only the top 1 workflow recommended by \textit{autoBagging}, on average, the kappa loss is half of the one we should expect from the suggestion made by the average rank method.

\begin{figure}[h]
	\centering
	\includegraphics[scale=0.65]{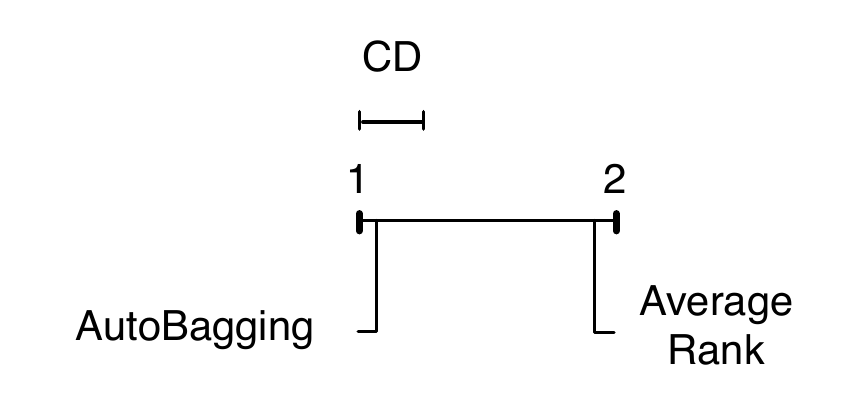}
	\caption{Critical Difference diagram (with $\alpha = 0.05$) of the experiments at the meta-level.}
	\label{cd_diagram_meta}
\end{figure}

\begin{figure}[h]
	\centering
	\includegraphics[scale=0.65]{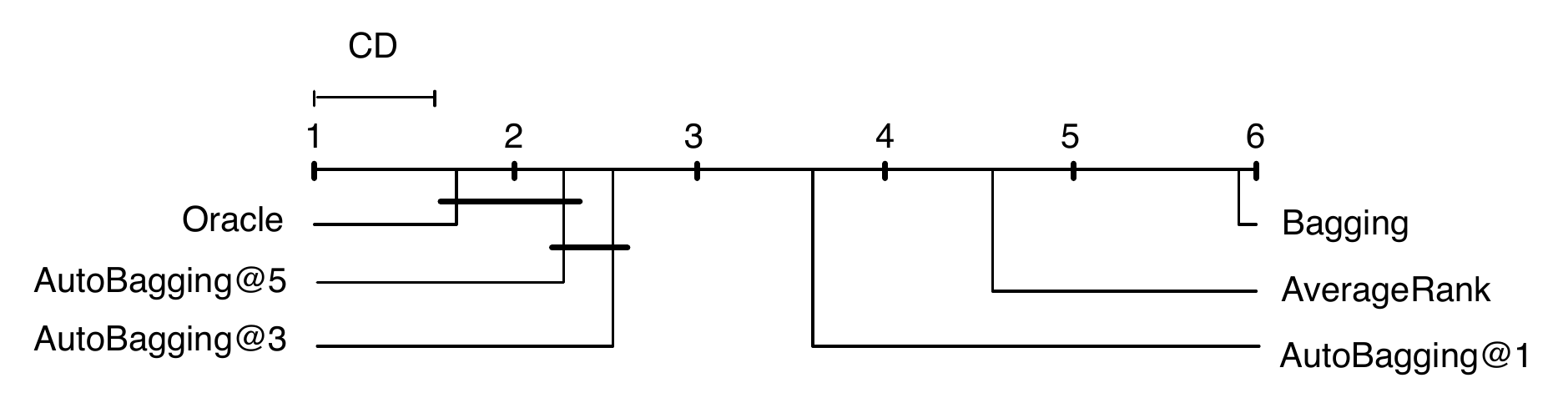}
	\caption{Critical Difference diagram (with $\alpha = 0.05$) of the experiments at the base-level.}
	\label{cd_diagram}
\end{figure}

We evaluated this results to assess their statistical significance using Dem{\v{s}}ar's methodology. Figures~\ref{cd_diagram_meta} and~\ref{cd_diagram} show the Critical Difference (CD) diagrams for both the meta and base-level.

At the meta-level, using MAP@10 as evaluation metric, \textit{autoBagging} presents a clearly superior performance in comparison with the Average Rank. The difference is statistically significant, as one can see in the CD diagram. This result is in accordance with performance that visualized in Figure~\ref{loss_curve} for both methods.

At the base-level, we compared \textit{autoBagging} with three baselines, as mentioned before: bagging with 100 decision trees, the Average Rank method and the oracle. We test three versions of \textit{autoBagging}, taking the top 1, 3 and 5 bagging workflows ranked by the meta-model. For instance, in \textit{autoBagging@3}, we test the top 3 bagging workflows ranked by the meta-model and we choose the best.

Starting by the tail of the CD diagram, both the Average Rank method and \textit{autoBagging@1} show a superior performance than Bagging with 100 decision trees. Furthermore, \textit{autoBagging@1} also shows a superior performance than the Average Rank method. This result confirms the indications that we visualized in Figure~\ref{loss_curve}.

The CD diagram shows also \textit{autoBagging@3} and \textit{autoBagging@5} have a similar performance. However, and we must highlight this results, \textit{autoBagging@5} shows a performance that is not statistically different from the oracle. This is extremely promising since it shows that the performance of \textit{autoBagging} excels if the user is able to test the top 5 bagging workflows ranked by the system.

\subsection{Discussion}

We decided to not include time in the experiments since \textit{autoBagging} execution time only depends on the computation of metafeatures. Given the nature of these metafeatures, such as entropy or mutual information, the computation is extremely fast (no more than a couple of minutes for the largest datasets used in the experiments).

\begin{figure}[h]
	\centering
	\includegraphics[scale=0.55]{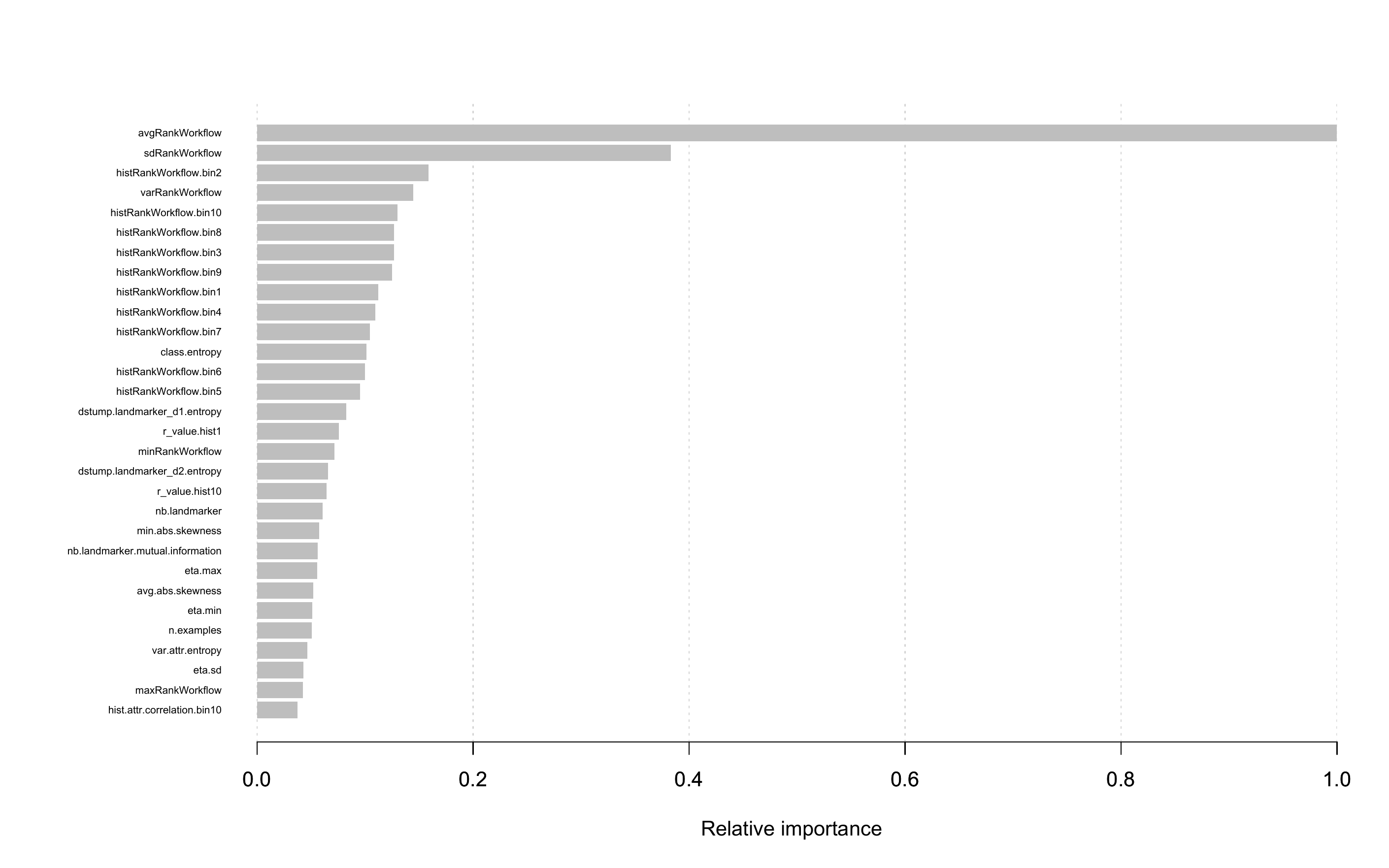}
	\caption{Top 30 most important metafeatures for the XGboost meta-model  measured using \textit{Gain}, which represents the relative contribution of the corresponding feature to the model calculated by taking each feature's contribution for each tree in the model.}
	\label{top_metafeatures}
\end{figure}

Figure~\ref{top_metafeatures} shows the relative importance of the top 30 most important metafeatures. It is clear that the most informative metafeatures are the ones generated using the rank of each workflow in the meta-training set as meta-function. Given that these metafeatures do not vary that much from dataset to dataset, we can assume that they are very important to characterize the bagging workflows. On the other hand, the remaining metafeatures are critical for the ability of the meta-model to generalize for all datasets. Metafeatures such as \textit{class.entropy},\textit{ dstump.landmarker\_d1.entropy} and \textit{r\_value.hist1} are also among the most informative metafeatures.

\section{Conclusion and Future Work}
\label{sec:cfw}

This paper presents \textit{autoBagging}, an \textit{autoML} system that makes use of a learning to rank approach and metalearning to automatically suggest a bagging ensemble specifically designed for a given dataset. We tested the approach on 140 classification datasets and the results show that \textit{autoBagging} is clearly better than the baselines to which was compared. In fact, if the top five workflows suggested by \textit{autoBagging} are tested, results show that the system achieves a performance that is not statistically different from the oracle, a method that systematically selects the best workflow for each dataset. For the purpose of reproducibility and generalizability, \textit{autoBagging} is publicly available as an R package.

As future work, we plan to further improve the experimental setup of \textit{autoBagging} by comparing it with state-of-the-art systems such as auto-sklearn and the hyperband method. Furthermore, we plan to study how we can use bayesian optimization to further improve the final ensemble, always taking into account concepts such as diversity and complementarity between models to design the final ensemble.

\vskip 0.2in
\bibliography{biblio}

\end{document}